\documentclass{article}
\usepackage{spconf,amsmath,graphicx}
\usepackage{subcaption}
\title{SIDU:~Similarity Difference and Uniqueness method for Explainable AI}
\name{Satya M. Muddamsetty$^{}$ \qquad Mohammad N. S. Jahromi$^{}$ \qquad Thomas B. Moeslund$^{}$}
 \address{$^{}$Visual Analysis of People Laboratory (VAP), Aalborg University, Aalborg, Denmark} 
\begin{document}
%\ninept
%
\maketitle
\begin{abstract}
% Since a deep learning algorithm can easily contain 50 million free parameters, it is nearly impossible to explicitly understand the inner workings of all details and hence explain exactly why an algorithms arrives at a certain conclusion or decision.
A new brand of technical artificial intelligence ( Explainable AI ) research has focused on trying to open up the 'black box' and provide some explainability. This paper presents a novel visual explanation method for  deep leaning networks in the form of a saliency map that can effectively localize entire object regions.  In contrast to the current state-of-the art methods, the proposed method shows quite promising visual explanations that can gain greater trust of human expert. Both quantitative and qualitative evaluations are carried out on both general and clinical data sets to confirm the effectiveness of the proposed method.  
\end{abstract}
\begin{keywords}
Explainable AI, Visual explanation, Clinical application, CNN
\end{keywords}
\vspace{-0.2cm}
\section{Introduction}
\label{sec:intro}
Deep Learning has resulted in a significant performance breakthrough in a wide variety of areas in computer vision tasks such as object detection \cite{girshick2014rich}, image classification \cite{oquab2015object}, image captioning \cite{fang2015captions} and many problems in machine learning (ML) \cite{antol2015vqa, das2018embodied, de2017guesswhat, lipton2018mythos}.  Despite all the popularity and high performance of deep learning, it is difficult to clearly understand or visualize the inner stacked layers of its architecture and ultimately, interpreting the output decision of such a network with millions of free parameters. Without principled understanding of how the 'black box' achieves its result, it is difficult to trust and deploy such AI models in, for example,  domain like medical diagnosis or criminal justice where, the final decision may have serious consequences. Therefore, Explainable AI (XAI) is a newly emerging discipline of AI that attempts to shed light on the ‘black box’ by providing visual explanation or analysis of feature
representations, hidden inside deep learning models. In this way, a particular deep learning model can be further evaluated by a human user/expert to establish trust in final predictions or help fix any classification errors. For example, in spite of a justifiably high skepticism rate within the medical community in supporting clinical decisions made by powerful machine learning models, XAI provides clinicians a useful tool by which to better audit the model’s predictions and reason about external factors influencing prediction such as bias in the data \cite{wu2019optimizing}.

Although there have been a number of early studies focusing on generating explanation schemes for deep network models, considering the complexity of such a challenging task there is still much more effort needed to establish both reliable quantitative and qualitative methods in this new field. The majority of the proposed  methods are based on generating visual feature explanations known as\textbf{ saliency maps}. For example, in \textit{Gradient-based algorithms} such as Grad-Cam \cite{GRAD-CAM}, DeCovNet \cite{bach2015pixel} or LRP~\cite{zeiler2014visualizing}, such a map can be obtained by backtracking the network’s activations from the output back to the input via backpropagation \cite{fong2019understanding}, in order to highlight the input regions that are most important in realizing the final prediction. Although the visual explanation of this class results in a well-detailed activation heat-map, computing the gradient for certain architecture models is not very straightforward \cite{zhou2016learning}.

\begin{figure*}
\begin{centering}
\includegraphics[width= 0.23\textwidth, height = 2.5cm]{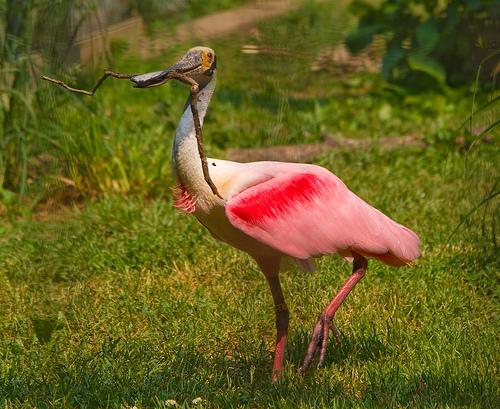}%results.eps}  
\hspace{0.01\columnwidth}
\includegraphics[width= 0.23\textwidth, height = 2.5cm]{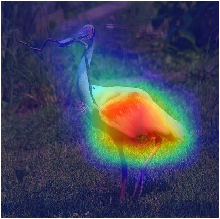}%results.eps}
\hspace{0.01\columnwidth}
\includegraphics[width= 0.23\textwidth, height = 2.5cm]{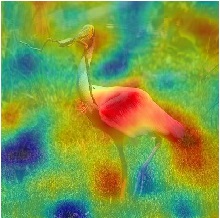}%results.eps}
\hspace{0.01\columnwidth}
\includegraphics[width= 0.23\textwidth, height = 2.5cm]{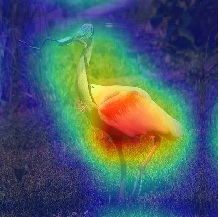}%results.eps}

\par\end{centering}
\begin{centering}
\vspace{1mm}
\par\end{centering}
\centering{}
\hspace{2mm} (a) Original image\hspace{2cm} b) GRAD-CAM\hspace{2.5cm} (c) RISE\hspace{2.5cm}(d) SIDU ( proposed~) \hspace{1.5cm}
\vspace{-.3cm}
\caption{An example of failure of saliency maps to capture entire object class in (b):GRAD-CAM and (c): RISE.}
\label{f11}
\end{figure*}

Alternatively, more recent approaches, like the state-of-the-art method of RISE \cite{RISE}, attempts to find the effect of selectively inserting or deleting parts of the input (\textit{Perturbation-based}) in the model's output prediction. Despite more accurate saliency maps and higher classification scores (after a selective deletion process of the input) of Perturbation-based methods compared to those of the gradient-based methods, it is not yet possible to visualize all the perturbations and determine which one characterizes the desired explanation best. Moreover, for both the gradient based and perturbation explanation methods, the generated visual explanation suffers from localizing the entire salient regions of an object required for higher classification scores. An example of this drawback is illustrated in Fig.~\ref{f11} where the estimated saliency maps were unable to localize entire object class properly. With more general interpretation approaches (\textit{Approximation-based}), as in the case of LIME \cite{ribeiro2016should} that creates its saliency map based on random superpixels, the problem becomes even more deteriorated. This issue plays a significant role in certain classification tasks such as those found in medical domain where generating highly accurate visual explanations are equivalent to highlighting complete regions of interest.\\
%%%%%%%%%%%%%%%%%%%%%%%%%%%%%%%%%%%%%%%%%%%%%%%%%%%%%%%%%%%%%%%%%%%%%%%%%%%%%%%%%%%%%%
% \begin{figure}[!hbt]
%     \centering
%     \begin{subfigure}[b]{0.31\linewidth} 
%         \centering
%         \includegraphics[width=1.2\textwidth,height = 2.5cm]{risefish.png}
%         \hspace{0.01\columnwidth}
%         \caption{RISE}
%         \label{fig:1-1}
%     \end{subfigure}
%       \begin{subfigure}[b]{0.31\linewidth}
%         \centering
%         \includegraphics[width=1.3\textwidth, height = 2.5cm]{Gradfish.png}
%         \hspace{0.01\columnwidth}
%         \caption{Grad-CAM}
%         \label{fig:1-2}
%     \end{subfigure}
%       \begin{subfigure}[b]{0.31\linewidth}
%         \centering
%         \includegraphics[width=1\textwidth,height =2.5cm]{Gradfish.png}
%         \hspace{0.01\columnwidth}
%         \caption{Grad-CAM}
%         \label{fig:1-2}
%     \end{subfigure}
%     \caption{An example of failure of saliency maps to capture entire ocject class.}
%     \label{fig:1} 
% \end{figure}

%%%%%%%%%%%%%%%%%%%%%%%%%%%%%%%%%%%%%%%%%%%%%%%%%%%%%%%%%%%%%%%%%%%%%%%%%%%%%%%%%%%%%%%%%%%%%
% \noindent With more general interpretation approaches (\textit{Approximation-based}), as in the case of LIME \cite{ribeiro2016should} that creates its saliency map based on random superpixels, the problem becomes even more deteriorated. This issue plays a significant role in certain classification tasks such as those found in medical domain where generating highly accurate visual explanations are equivalent to highlighting complete regions of interest.\\
%%%% Say a sentence on whit box method
\vspace{-.25cm}
In this paper, motivated by studies in both GRAD-CAM and RISE, we propose a new visual explanation approach for estimating pixel saliency by extracting the last convolutional layer of the deep CNN model and creating the similarity difference mask which is eventually combined to form a final map for generating the visual explanation for the prediction. We refer to the proposed method as \textit{similarity difference and uniqueness method (SIDU)}. The SIDU method is gradient-free (as opposed to GRAD-CAM) and unlike the random mask mechanism of the RISE algorithm, the final combined mask of our proposed methods comes from the last activation masks of the CNN model. The algorithm provides much better localization of the object class in question (see, for example, Fig.~\ref{f11}~(d)). This results in gaining greater trust of human expert level to rely on the deep model. Quantitative and qualitative tests on both general and clinical datasets further demonstrate the effectiveness of our proposed approach compared to the state-of-the-art RISE method. 

\begin{figure}[h]
 \includegraphics[width=10cm,height=10cm]{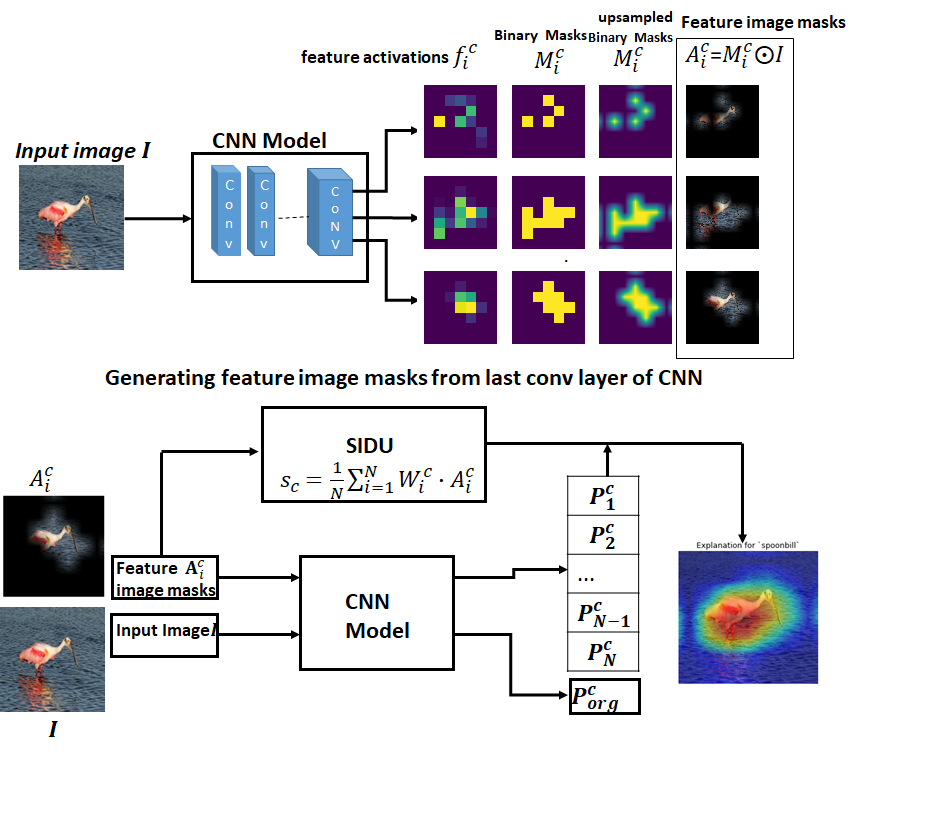}
 \caption{Block diagram of the proposed explanation method.}
\label{f22}
 \end{figure}
 \vspace{-0.5cm}
\section{PROPOSED METHOD}
\label{sec:proposed_method}
Overview of the proposed explanation method is illustrated in Fig.~\ref{f22} In the following subsections, we describe each step.
%In this section, we introduce our proposed explainable model for the deep learning methods.
%Our method is composed of three steps, First we extract the last convolution layer of the CNN to generate the mask using the last convolution layer of the given model. Second, we compute the similarity differences for each mask with respect to predicted class and Finally we compute the weights of each masks we combine them into final map which shows the explanation of the prediction. The steps involved in proposed methods are described in following subsections.
%An overview of the method is presented in Fig. 1, and the different steps are described in the following subsections.
\vspace{-.3cm}
\subsection{Generating Masks}
We first generate masks from the last convolution layers of the deep CNN model. Let us consider any deep CNN model $\textit{F}$ with last convolution layers of size $n\times n \times N$ where $'n'$ is size of that layer and $'N'$ is the number of features in the activation maps $\textbf{f}$ of class $c$, i.e., $\textbf{f}^{c} = [f^{c}_{i},....f^{c}_{N}]$. Each feature activation map $f^{c}_{i}$ is then converted into a binary mask $M^{c}_{i}$ corresponding to the feature activation map $f^{c}_{i}$ in the convolution layer. Next, a bi-linear interpolation is applied to up-sample the binary mask for a given input image $I$ of size $W \times H$. After interpolation, the binary mask $M^{c}_{i}$ will be no longer binary and the values range between $[0,1]$. Point-wise multiplication is performed between interpolated binary mask $M^{c}_{i}$ and input image $I$ and is represented as
\begin{equation} 
    A^{c}_{i} = F(I \odot M^{c}_{i})
\end{equation}
where $F$ is an CNN model, $A^{c}_{i}$ is the feature activation image mask of feature map $f^{c}_{i}$ and $i= 1,....N$.
%The scalar confidence score is predicted for the given input image $I$ by the CNN model $F(I\odot M_{i})$. 
\vspace{-.25cm}
\subsection{Computing Similarity Differences and Uniqueness}
%Once the binary masks are generated and point-wise multiplication is performed between binary mask $M^{c}_{i}$ and Image $I$,%
\vspace{-.2cm}
In the next step, we compute probability prediction scores for all the feature activation image masks $\textbf{A}^{c}$ of class $c$, i.e., $\textbf{A}^{c}= [A^{c}_{i},....A^{c}_{N}]$ using the CNN model $F$, respectively. Let the probability prediction score of the feature activation image mask $A^{c}_{i}$ be $P^{c}_{i}$ and the probability prediction score for the original image $I$ be $P^{c}_{org}$. The similarity differences are then computed between each input feature activation image mask prediction score $P^{c}_{i}$ and prediction score $P^{c}_{org}$ of the original image $I$. The main idea is that the relevance of a feature map is estimated by measuring how the prediction changes if the feature is unknown, $i.e.,$ the similarity difference between prediction scores. The relevance value of the feature  activation image mask increases if it is similar to the predicted class and decreases otherwise. The similarity difference of set of feature activation maps $SD^{c}_{i}$ is given by
\begin{equation}
 SD^{c}_{i} = \exp{(\frac{-1}{2\sigma^2}\|P^{c}_{org}-P^{c}_{i}\|)}
\end{equation}
%where ${P}^{c}_{org}$ and ${P}^{c}_{i}$ are the predictions for the original input image and $i^{th}$ masked image generated from the last convolution layers.
We also compute a uniqueness measure $U$ which implements the commonly employed assumption that image regions which stand out from other regions in certain aspects catch our attention and hence should be labeled more salient. We therefore evaluate how different each respective feature mask is from all other feature masks constituting an image. The intuition behind this is to suppress the false regions with low weights and highlight the actual regions which are responsible for predictions with higher weights.
The uniqueness measure $U^{c}_{i}$ is given in Eq.\ref{eq:uniqness}. 
\begin{equation} \label{eq:uniqness}
    U^{c}_{i} = \sum_{j}^{N}\|P^{c}_{i}-P^{c}_{j}\|,
\end{equation}
%\vspace{-.2cm}
The final weight of feature importance $W^{c}_{i}$ is the dot product of the similarity difference $SD^{c}_{i}$ and uniqueness measure $U^{c}_{i}$ and is given
\begin{equation} \label{eq:sim_uniqness}
    W^{c}_{i} = SD^{c}_{i}\cdot U^{c}_{i},
\end{equation}
where $SD^{c}_{i},U^{c}_{i}$ are the similarity difference and uniqueness values for the feature activation image mask $A^{c}_{i}$ of the object class $c$.

\vspace{-.3cm}
\subsection{Explanations for the prediction}
The final visual explanation map $S_{c}$, also known as the class discriminative localization map can be computed as a weighted sum of image masks $A^{c}_{i}$ , where the weights are computed by Eq.~\ref{eq:sim_uniqness}. Thus the visual explanation of the predicted class $c$ is given by
\vspace{-.2cm}
\begin{equation} \label{eq: visual_exp}
    S_{c} = \frac{1}{N}\sum_{i}^{N} W^{c}_{i} \cdot A^{c}_{i}
\end{equation}
In summary, to explain the decision of the predicted class $c$ visually, we first extract the last convolution layer from the deep CNN model $F$ which has $N$ number of features activation maps of size $n \times n$. We then generate $N$  binary masks and point wise multiplication is performed between each generated binary mask $M_{i}$ and the input image $I$. The similarity difference $SD$ between probability scores of the predicted class and each point-wise multiplied image mask $A_{i}$ and uniqueness measure $U$ between the image masks are computed. Weights $W_{i}$ of each image mask $A_{i}$ computed by the dot product of $SD$ and $U$. Finally, the visual explanation $S_{c}$ is a weighted sum of  feature activation image masks $A_{i}$ given in Eq.~\ref{eq: visual_exp}

\begin{figure*}[hbt!p]
\begin{centering}
\includegraphics[width= 0.15\textwidth, height = 2.5cm]{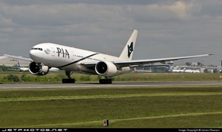}%results.eps}  
\hspace{0.01\columnwidth}
\includegraphics[width= 0.15\textwidth, height = 2.5cm]{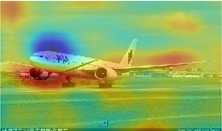}%results.eps}
\hspace{0.01\columnwidth}
\includegraphics[width= 0.15\textwidth, height = 2.5cm]{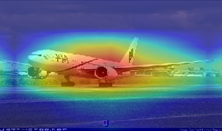}%results.eps}
\hspace{0.01\columnwidth}
\includegraphics[width= 0.15\textwidth, height = 2.5cm]{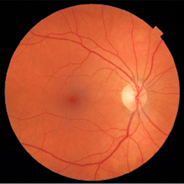}%results.eps}
\hspace{0.01\columnwidth}
\includegraphics[width= 0.15\textwidth, height = 2.5cm]{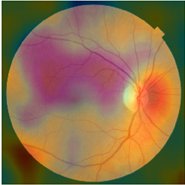}%results.eps}
\hspace{0.01\columnwidth}
\includegraphics[width= 0.15\textwidth, height = 2.5cm]{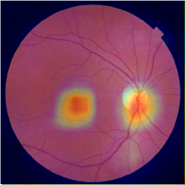}%results.eps}
\par\end{centering}
\begin{centering}
\vspace{1mm}
 
\par\end{centering}
\begin{centering}
\includegraphics[width= 0.15\textwidth, height = 2.5cm]{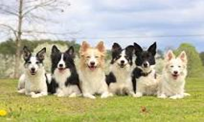}%results.eps}  
\hspace{0.01\columnwidth}
\includegraphics[width= 0.15\textwidth, height = 2.5cm]{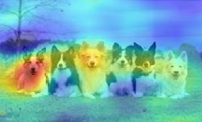}%results.eps}
\hspace{0.01\columnwidth}
\includegraphics[width= 0.15\textwidth, height = 2.5cm]{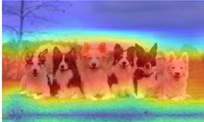}%results.eps}
\hspace{0.01\columnwidth}
\includegraphics[width= 0.15\textwidth, height = 2.5cm]{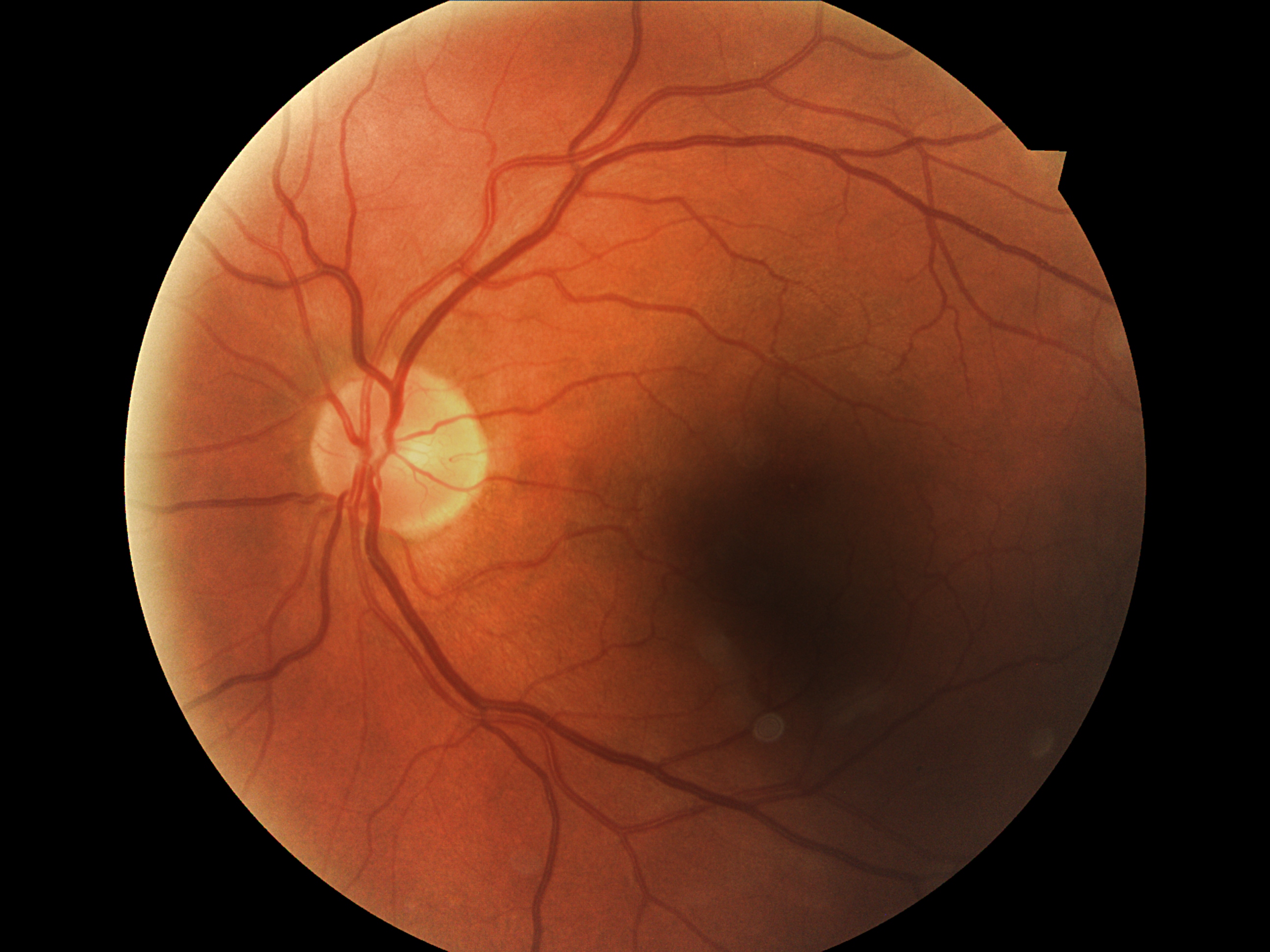}%results.eps}
\hspace{0.01\columnwidth}
\includegraphics[width= 0.15\textwidth, height = 2.5cm]{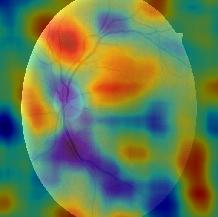}%results.eps}
\hspace{0.01\columnwidth}
\includegraphics[width= 0.15\textwidth, height = 2.5cm]{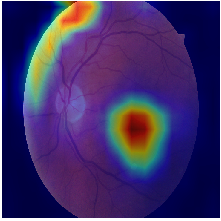}%results.eps}
\par\end{centering}
\centering{}
\hspace{-1.2cm}(a) Original Image \hspace{0.85cm}(b) RISE\hspace{1.6cm}(c) SIDU\hspace{1cm}(d) Original Image\hspace{0.7cm}(e) RISE\hspace{1.6cm}(f) SIDU
% \caption{Examples of different quality grades of  RFIQA dataset.(a) Grade0, (b) Grade1, (c) Grade2, (d) Grade3, (e) Grade4 (f) Grade5  \label{fig:RFIQA dataset}}
\caption{Saliency maps generated for the natural images class of ImageNet~~$<$(a), (b), (c)$>$~~ and Good~/~Bad quality eye fundus images ~~$<$(d), (e), (f)$>$~~ from  RFIQA by RISE and the SIDU method with ResNet50 as the base network. In practice,~the doctors verify the visibility of the optical disc and macular regions in a good quality image ($4^{th}$ image, $1^{st}$ row) corresponding to the highlighted regions in the heatmap of the proposed method. Similarly, the bad quality ($4^{th}$ image, $2^{nd}$ row) is due to the shadow just above the center of the image,~i.e., exactly the region highlighted by the proposed method.} 
\label{f1s}
\end{figure*}

\section{Experiments and Results}
\label{sec:Exp_results}
%In this section we evaluate the performance of the proposed visual explanation method. 
A good explanation must be consistent with the CNN model's prediction and visualize the results in a manner that is intuitive for a human. In order to evaluate the performance of the SIDU explanation method we choose two datasets with different characteristics. The ImageNet~\cite{ILSVRC15} dataset of Natural Images with 1000 classes. We use 2000 images randomly collected from the ImageNet validation dataset. The other is a  Retinal Fundus Image Quality Assessment (RFIQA) dataset from the medical domain. The dataset consistes of 9,945 images with two levels of quality,  'Good' and 'Bad'. The retinal images were collected from a large number of patients with retinal diseases~\cite{sat20}.
%The RFIQA dataset composed of 9945 images with 'Good' and 'Bad' levels grades. 
%The retinal images in RFIQA dataset have different characteristics collected from large number of patients with retinal diseases. 

\noindent There is no consensus about what interpretability is in machine learning. Nor it is clear how to measure it. Initial research attempts to formulate approaches for evaluation~\cite{poursabzi2018manipulating} which suggest faithfulness and human trust. We first evaluate the faithfulness of our model by studying the correlation between the visual explanation and prediction. Second, an expert level evaluation is performed, where human domain experts are involved in evaluating the human interpretability or human trust of the model. The experimental evaluation of faithfulness and human trust of the SIDU model is described in Section~\ref{sec:faitfulness} and Section~\ref{sec:qualitative}, respectively.
\vspace{-0.3cm}
\begin{table}[tb!]
\caption{Quantiative Comparision XAI methods for Resent50 on ImageNet validation set.}\label{t1}
\begin{centering}
\begin{tabular}{|c|c|c|}
\hline 
METHODS & Insertion$\uparrow$ & Deletion$\downarrow$  \tabularnewline
\hline 
\hline 
RISE~\cite{RISE} & 0.63571 & 0.13505   \tabularnewline
\hline 
SIDU & \textbf{0.65801} & \textbf{0.13424}   \tabularnewline
\hline
\end{tabular}
\par\end{centering}
\end{table}

\begin{table} [t!]
\caption{Quantiative Comparision XAI methods on RFIQA dataset for our trained ResNet-50 model. }\label{t2}
\begin{centering}
\begin{tabular}{|c|c|c|}
\hline 
METHODS & Insertion$\uparrow$ & Deletion$\downarrow$  \tabularnewline
\hline 
\hline 
RISE~\cite{RISE} & 0.75231 & 0.59632 \tabularnewline
\hline 
SIDU & \textbf{0.87883} & \textbf{0.47818}   \tabularnewline
\hline
\end{tabular}
\par\end{centering}
\end{table}
\subsection{Evaluating Faithfulness} \label{sec:faitfulness}
The faithfulness of the proposed method is evaluated using two automatic causal metrics \textit{insertion} and \textit{deletion} proposed by~\cite{RISE}. We choose these metrics to compare with the state-of-art methods. The \textit{deletion metric} removes the saliency region responsible for predicting the object class in the image and forces the base model to change its decision. As more pixels are removed from the saliency region, this metric measures a decrease in the probability of the predicted score. Good explanation shows a sharp drop in the predicted score and thus a low area under the probability curve. On the other hand the \textit{insertion metric} is a complementary approach. It measures the increase in probability of predicted score as more and more pixels are included with higher Area Under Curve (AUC) as an indication of a good explanation.

\noindent To perform the experimental evaluation of the proposed explanation method we conduct two experiments. In the first experiment we choose one of the existing standard CNN models, ResNet-50~\cite{Resnet-50} pre-trained on ImageNet dataset~\cite{ILSVRC15} and evaluate the faithfulness of the explainable model on the ImageNet validation dataset. Table \ref{t1} summarizes the results obtained by the proposed method and compares it to most recent work~\cite{RISE}. We can observe that the proposed method achieved better performance for both metrics, i.e.,  outperforming RISE~\cite{RISE}. This can be explained by the fact that the RISE method generates $N$ number of random masks and the weights predicted for these masks give higher weights to false regions which makes the final map of RISE  noisy. An example is shown in Fig.~\ref{f1s}. In our proposed method, however, the generated masks come from the last feature activation maps of the CNN model. Due to this the final explanation map will localize the entire region of interest (object class). The visual comparison of the explanations of the proposed method on ImageNet are shown in Fig.~\ref{f1s}~(a),~(b),~(c).
%\vspace{-.6cm}

\noindent To justify the above statement, we conducted a second experiment. We trained the existing ResNet-50~\cite{Resnet-50} with an additional two FC layers and softmax layer on the RFIQA dataset. The CNN model achieves $94 \%$ accuaracy. The proposed explanation method uses the trained model for explaining the prediction of the RFIQA  test subset with 1028 images. The evaluated results are summarized in Table~\ref{t2}. We can observe that the proposed method achieves higher AUC for \textit{insertion} and lower AUC for \textit{deletion} compared to RISE~\cite{RISE}. The visual explanations of the proposed and the RISE methods on the RFIQA test dataset are shown in Figure.~\ref{f1s}~(d),~(e),~(f).
\vspace{-.3cm}
%============================================================================================
%\begin{figure*}
 %   \centering
  %   \includegraphics[width=16cm,height=8cm]{Natural.PNG}
   % \caption{ }\label{f1s}
%\end{figure*}
%============================================================
%================================
\subsection{Qualitative Evaluation} \label{sec:qualitative}
Human expert level is an essential criterion for those end-users who have less trust in the results of prediction models (e.g. clinician). As in our medical diagnosis case, to show the effectiveness of the proposed method in terms of capturing the correct region with respect to the state-of-art method, we ask two ophthalmologist experts at the hospital to evaluate which visual representation invokes more trust and hence matches with actual examination results performed in the clinic. Here, the generated heat-maps in the RISE algorithm were treated as the baseline for the comparison.\\
Next, we follow the same setting as discussed in \cite{GRAD-CAM} and generate explanation heat-maps of 100 fundus images for two classes of ‘Good’ and ‘Bad’ quality using both the proposed method and the RISE algorithm. The exact nature of each algorithm in the test remains unknown to the ophthalmologists. Indeed, they are labeled as either 'model I' or 'model II' to the test participants. Once the ophthalmologist determined which model better represents the regions of interest (good/bad quality regions) for each image, we next calculate the relative frequency of each outcome per total fundus image. Note that each participant had the option to select “both” models if they feel both the generated explanation maps were rather similar. In such cases we may have three different possibilities for each test image. In the case of the first ophthalmologist, the RISE explanation map was selected with the relative frequency of $0.02$, the proposed algorithm with $0.84$ and $0.14$ being the same. For the second ophthalmologist, these numbers are  $0.05$, $0.93$ and $0.02$, respectively. This demonstrates that both ophthalmologists significantly favor the visual explanation generated by the proposed method over the RISE method. This can be clearly observed by visual examples of these explanation maps for the fundus image in Fig.~\ref{f1s}. It is visually evident that the proposed algorithm is capable of properly localizing the region of interest and hence gaining greater trust by the expert.
\vspace{-0.3cm}
\section{Conclusion}
\label{sec:conclusion}
 In this paper we proposed a novel method called SIDU for explanation of  black box models in a heat-map form via feature maps of the last convolution layers in the model. The proposed method is a gradient-independent method that can effectively localize entire object classes in an image. The quantitative and qualitative (human trust) experiments show that for both  general and critical medical data,  the proposed method outperforms state-of-the-art. The new explanation approach can provide further insight and helps in gaining greater trust in ML-based prediction results for the end-user in a sensitive-domain.    

\vfill\pagebreak

\bibliographystyle{IEEEbib}
\bibliography{refs}

\end{document}